\documentclass[10pt,twocolumn,letterpaper]{article}

\usepackage{wacv}
\usepackage{times}
\usepackage{epsfig}
\usepackage{graphicx}
\usepackage{amsmath}
\usepackage{amssymb}

\graphicspath{{figs/}}

\def\wacvPaperID{757} 

\wacvfinalcopy 
\def\assignedStartPage{9876} 

\usepackage[breaklinks=true,bookmarks=false]{hyperref}
\begin{document}

\title{ Overcomplete Deep Subspace Clustering Networks}

\author{Jeya Maria Jose Valanarasu  \hspace{2.5cm} Vishal M. Patel\\
	Department of Electrical and Computer Engineering,\\
Johns Hopkins University, 3400 N. Charles St, Baltimore, MD 21218, USA\\
{\tt\small \{jvalana1, vpatel36\}@jhu.edu}
}

\maketitle

\begin{abstract}
   Deep Subspace Clustering Networks (DSC) provide an efficient solution to the problem of unsupervised subspace clustering by using an undercomplete deep auto-encoder with a fully-connected layer to exploit the self expressiveness property. This method uses undercomplete representations of the input data which makes it not so robust and more dependent on pre-training. To overcome this, we propose a simple yet efficient alternative method - Overcomplete Deep Subspace Clustering Networks (ODSC) where we use  overcomplete representations for subspace clustering. In our proposed method, we fuse the features from both undercomplete and overcomplete auto-encoder networks before passing them through the self-expressive layer thus enabling us to extract a more meaningful and robust representation of the input data for clustering. Experimental results on four benchmark datasets show the effectiveness of the proposed method over  DSC and other clustering methods in terms of clustering error. Our method is also not as dependent as DSC is on where pre-training should be stopped to get the best performance and is also more robust to noise. Code - \href{https://github.com/jeya-maria-jose/Overcomplete-Deep-Subspace-Clustering}{https://github.com/jeya-maria-jose/Overcomplete-Deep-Subspace-Clustering}
   
\end{abstract}

\section{Introduction}

Subspace Clustering \cite{vidal2011subspace} is a learning paradigm which involves grouping a set of similar data points in an unsupervised way. Let  $X \in \mathbb{R}^{D \times N}$ be a matrix such that its columns are chosen from a union of $k$ subspaces of $\mathbb{R}^{D}$, $\cup_{i=1}^k \{S_i\}$ of dimensions $d_i$ where $d_i<<min\{D,N\}$, then the task of subspace clustering is to categorize the columns of $X$ into their corresponding subspaces. Subspace clustering is widely popular for its use in image clustering \cite{elhamifar2009sparse,yang2006clustering}, image segmentation \cite{yang2008unsupervised,ma2007segmentation}, motion segmentation \cite{elhamifar2013sparse,kanatani2001motion}, face clustering \cite{ho2003clustering} and various other computer vision tasks. As most of the data in real-world are high dimensional, there exists a requirement to convert the high dimensional data into meaningful subspace representations before clustering. A union of multiple subspaces can then be clustered together into a single category. The use of multiple subspaces is what differentiates subspace clustering from principal component analysis (PCA) related methods where it is assumed that data is drawn from a single low-dimensional subspace. 

Several methods have been proposed in the literature for solving the problem of subspace clustering  using conventional methods \cite{elhamifar2013sparse,you2016oracle,liu2010robust,you2016scalable,favaro2011closed,lu2013correlation,lu2012robust, LatentSSC_JSTSP, DA_SSC_BMVC, Mahdi_multimodal_fusion}. The first deep learning-based solution to the problem of subspace clustering, called Deep Subspace Clustering Networks (DSC), was introduced in \cite{ji2017deep}. DSC achieved a huge boost in performance when compared to previous conventional methods. DSC uses a convolutional autoencoder to learn a non-linear representation of the data. The network has an undercomplete (``encoder-decoder") architecture where the encoder gets trained to learn an abstract low-dimensional representation of the input image while the decoder learns to reconstruct the original image from those representations. This deep representation learned by the encoder explicitly performs a non-linear mapping of the data which helps in performing better subspace clustering. Although DSC performs very well and has a huge margin of improvement in terms of performance over the previous methods, there exists two main issues with the method. This method is not so robust and its performance drops considerably when there are potential degradations (i.e. noise) in the data leading to noisy representations \cite{zhou2018deep}. Another major issue with DSC is that it depends a lot on where pre-training is stopped. Even if the pre-training is stopped some epochs before or after the correct epoch where clear reconstructions start to appear, the clustering becomes unstable and the performance drops significantly. 


In this paper, we propose an alternative solution which improves the performance while being able to obtain a robust representation and a stable training. We propose Overcomplete Deep Subspace Clustering Networks (ODSC) where we make use of overcomplete representations which has greater robustness in the presence of noise and has a flexibility to match structures in the data. We induce overcomplete representations here by introducing an overcomplete convolutional autoencoder which is trained in parallel to the undercomplete autoencoder as in DSC. We then combine both the representations and use a self-expressive layer to learn pairwise affinities between the data points. This simple trick of fusing both overcomplete and undercomplete representations make the training stable and not be over-dependent on pre-training. We extensively analyze this issue by conducting various experiments.   The pre-trained ODSC autoencoder is also able to obtain far better reconstructions compared to the pre-trained DSC autoencoder  which only shows that better representations are learned by the overcomplete network. Even while maintaining the number of parameters to be the same as that of DSC, we get a good improvement over the performance of DSC with added advantages of more robustness and  stable training.  We evaluate our method on four different benchmark datasets: MNIST \cite{lecun1998gradient}, COIL20 \cite{nene1996columbia}, ORL \cite{samaria1994parameterisation} and Extended Yale B \cite{lee2005acquiring}. Our experiments demonstrate that ODSC significantly outperforms DSC and other conventional subspace clustering methods by a large margin.       

\section{Related Work}
Subspace clustering was initially solved by methods which relied on linear methods.  These methods first construct an affinity matrix by measuring the affinity for every pair of data points. Then methods like NCut (normalized cuts) \cite{shi2000normalized} or spectral clustering \cite{ng2002spectral} were used on the affinity matrix. These two problems are either solved sequentially \cite{elhamifar2009sparse,favaro2011closed,liu2010robust,lu2013correlation,lu2012robust} or in multiple passes in an alternate manner \cite{feng2014robust,guo2015robust,li2015structured,li2017structured,zhang2016low}. An affinity matrix can be built by exploiting the self-expressiveness property of data  \cite{elhamifar2009sparse,feng2014robust,li2015structured,liu2010robust,lu2012robust,ji2014efficient}, using  factorization methods \cite{costeira1998multibody,ji2015shape,kanatani2001motion,mo2012semi,vidal2008multiframe} or by using high-order model-based methods \cite{chen2009spectral,ochs2012higher,purkait2016clustering}. Out of these, self-expressiveness property-based methods are more robust to corruption by noise. The property of self-expressiveness corresponds to the ability to represent data points as a linear combination of other points in the same subspace.  Self-expressiveness can be formulated as follows: Given a set of data points $\{x_i\}_{i=1,...,N}$ which are taken from a collection of linear subspaces $\{S_i\}_{i=1,...,N}$, define a matrix $X$ whose columns are stacked up with $x_i$. The self-expressiveness property can then be represented as $X=XC$,  where $C$ is the self-representation matrix. It has been shown that if the subspaces are independent, then $C$ is guaranteed to have a block diagonal structure \cite{ji2014efficient}. This means that if points $x_i$ and $x_j$ lie in the same subspace then the corresponding coefficient $c_{ij}$ in matrix $C$ cannot be zero. To build an affinity matrix for spectral clustering, matrix $C$ can be used. This idea can be mathematically represented as an optimization problem as follows:

\begin{equation}
\min_C \|C\|_p + \frac{\lambda}{2} \|X-XC\|_F^2 \; \; \; s.t. \; \; \; (\text{diag}(C)=0),
\end{equation}

where $p= 0, 1$ or nuclear norm. The diagonal constraint here avoids trivial solutions for sparsity inducing norms. A major drawback of this clustering approach is that this holds true only for linear subspaces. For solving cases where data points do not lie in the linear subspace, several non-linear kernel-based methods have also been proposed \cite{patel2013latent,patel2014kernel,xiao2015robust,yin2016kernel, LatentSSC_JSTSP} which require a pre-defined kernel. However, these predefined kernels cannot be assured to provide feature spaces that are well suited for subspace clustering.

Following the popularity of deep learning methods in various tasks of computer vision and machine learning like image segmentation, image restoration, medical image analysis, etc., deep learning was explored for subspace clustering in DSC \cite{ji2017deep}. DSC also introduced a novel self-expressive layer for deep autoencoders so as to train an autoencoder such that its latent representation is well-suited for subspace clustering. This idea of harnessing the self-expressiveness property using a deep autoencoder resulted in a huge boost in performance when compared to other conventional methods \cite{Mahdi_nips2020}. Based on this work, several extensions were proposed very recently exploring  adversarial learning \cite{zhou2018deep, DA_ISBA_2018}, multimodal inputs \cite{abavisani2018deep}, multi-level representations \cite{kheirandishfard2020multi},  and self-supervised learning \cite{zhang2019self} for subspace clustering. Other spectral clustering free techniques like distribution preserving DSC \cite{zhou2019latent} and \cite{zhang2019neural} have also been proposed. More recently, works like Deep Latent Low-Rank Fusion Network \cite{zhang2020deep},  Block-diagonal Adaptive representation \cite{zhang2019robust}, Multilayer Collaborative Low-Rank Coding Network \cite{li2019multilayer} have also been proposed for subspace clustering.   In contrast to these methods, we focus completely on DSC and show how it can be easily  made very efficient and robust using overcomplete representations.

\section{Overcomplete Deep Subspace Clustering Networks (ODSC)}

The proposed approach makes use of overcomplete representations to improve the clustering performance. In this section, we first briefly describe the concept of overcomplete representations before explaining our proposed network architecture, clustering method and training strategy. 

\subsection{Overcomplete Representations}

Overcomplete Representations \cite{lewicki2000learning} were first introduced as an alternative and a more general method for signal representation. It involved using overcomplete bases (overcomplete dictionaries) so that the number of basis functions is more than the number of input signal samples.  This enables a higher flexibility for capturing structure in data and thus it is shown to be more robust. From \cite{vincent2008extracting}, we can see that overcomplete auto-encoders acted as better feature extractors for denoising. Interestingly, the idea of overcomplete representations have been very much under-explored in deep learning. All the major architectures widely used in deep learning use a generic ``encoder-decoder" architecture in which the encoder tries to extract an abstract version of the input data and the decoder learns to take the latent low-dimensional representation back to a high-dimensional output depending on the task at hand. This generic ``encoder-decoder" model is an example of undercomplete representations as the number of spatial dimensions is less in the latent space when compared to the input data.  This is accomplished in a deep convolutional undercomplete auto-encoder where the convolutional layers are followed by max-pooling layers in the encoder and by upsampling layer in the decoder. Max-pooling reduces the spatial dimensionality of the feature maps while upsampling does the opposite. Note that this arrangement of an undercomplete network's encoder is what forces the initial layers of a deep network to learn low-level features and the deep layers to learn high-level features. This happens because the receptive field of the filters increases after every max-pooling layer. With an increased receptive field, the filters in the deep layers have access to more pixels in the initial image thus enabling them to learn high-level features. Recently, overcomplete representations have been explored for medical image segmentation \cite{valanarasu2020kiu, valanarasu2020kiunet} and image-deraining \cite{yasarla2020exploring}. 

   In this work, we propose using an overcomplete deep autoencoder, where the encoder takes the input data to a higher spatial dimension. This is achieved by using an upsampling layer after every convolutional layer in  the encoder. Note that, the dimensionality  of the latent representations for any convolutional network depends on the  number of filters and the feature map size used in a network. In  \cite{vincent2008extracting}, an overcomplete fully-connected network is defined as a network which has more number of neurons for representation in its hidden layers than in the initial layers.  Similarly in this paper, we define an overcomplete CNN as a network that takes the input to a higher dimension in its deeper layers (spatially).

 \begin{figure}[htbp!]
 	\begin{center}
 		\centering
 		\includegraphics[width=0.45\textwidth]{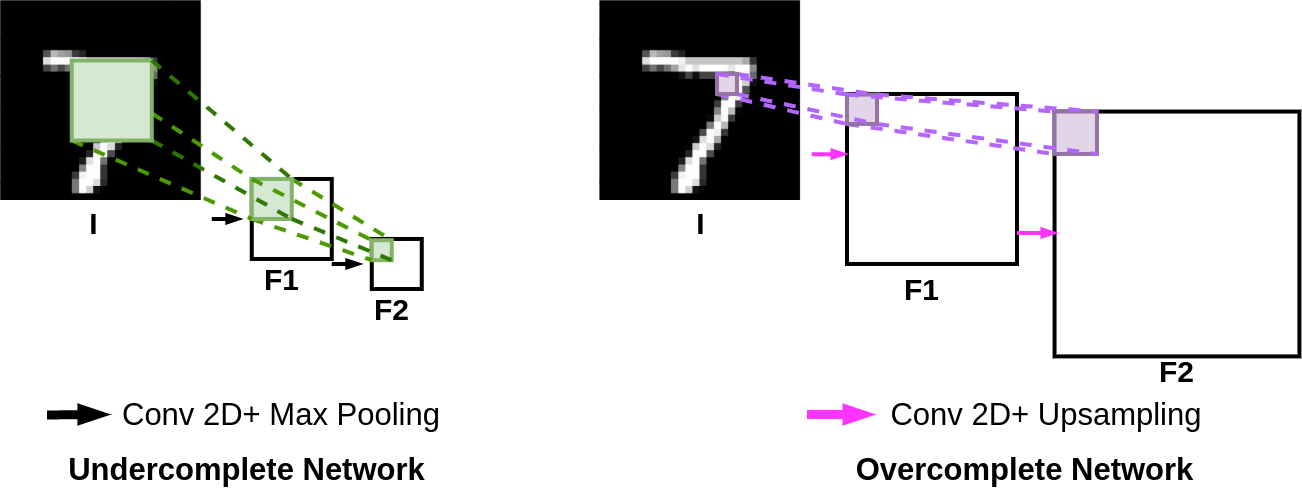}
 		\\
 		(a)\hskip125pt(b)\hskip125pt
 		\caption{Explanation of how receptive field changes in an (a) undercomplete network architecture, and in an (b) overcomplete network architecture. }
 		\label{Fig:expl}
 	\end{center}
 \end{figure}
 
    Replacing max-pooling layers with upsampling layers in the encoder causes the receptive field size to be constrained  in the deeper layers making the deeper layers learn more fine details than the initial layers as seen in Fig \ref{Fig:expl}. To understand this further, let $I$ be the input image, $F_1$ and $F_2$ be the feature maps extracted from the conv layers 1 and 2, respectively. The max-pooling layer present in these conv layers of the undercomplete architecture (as seen in Fig \ref{Fig:expl}(a)) is the main reason why the receptive field is large in the successive layers. Let the initial receptive field of the conv filter be $k \times k$ on the image. The receptive field size change due to max-pooling layer is dependent on two variables- pooling coefficient and stride of the pooling filter. For convenience, the pooling coefficient and stride is both set as 2 (as in most of the networks). Considering this configuration, the receptive field of conv layer 2 (to which $F_1$ is forwarded) on the input image would be $ 2 \times k \times 2 \times k$. Similarly, the receptive field of conv layer 3 (to which $F_2$ is forwarded) would be $ 4 \times k \times 4 \times k$. This increase in receptive field can be generalized for the $i^{th}$ layer in an undercomplete network as follows:
    
    \[ RF (w.r.t \; I) =  2^{2*(i-1)} \times k \times k. \]
    
    For the proposed overcomplete network, we have upsampling layer of coefficient 2  replacing the max-pooling layer. As the upsampling layer actually works exactly opposite to that of max-pooling layer, the receptive field of conv layer 2 on the input image now would be $ \frac{1}{2} \times k \times \frac{1}{2} \times k$. Similarly, the receptive field of conv layer 3  now would be $ \frac{1}{4} \times k \times \frac{1}{4} \times k$. This increase in receptive field can be generalized for the $i^{th}$ layer in the overcomplete branch as follows:  
    
    \[ RF (w.r.t \; I) =  \left(\frac{1}{2}\right)^{2*(i-1)} \times k \times k. \] 
    
\begin{figure*}[htbp!]
	\begin{center}
		\centering
		\includegraphics[width=0.8\textwidth]{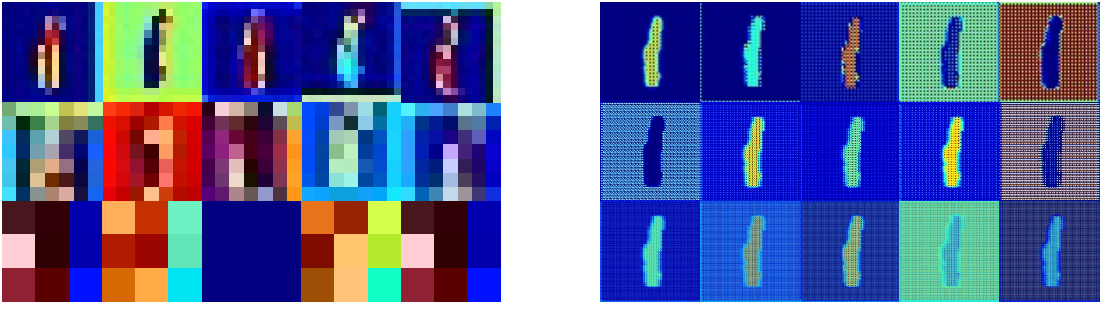}
		\\
		(a)\hskip210pt(b)\hskip125pt
			\caption{Feature maps captured using (a) Undercomplete network architecture. (b) Overcomplete network architecture. The rows represent the layer from which the feature maps were taken. Row 1 corresponds to layer 1, Row 2 corresponds to layer 2 and Row 3 corresponds to layer 3.}
		\label{Fig:feat}
	\end{center}
\end{figure*}

    This helps in the overcomplete network learn more low-level information like edges and other finer details better. In Figure \ref{Fig:feat}, we visualize some of the feature maps learned by the  undercomplete and overcomplete networks while trained on the MNIST dataset. We can observe that the learned features in the overcomplete network are more detailed and capture the edges perfectly when compared to the features extracted from the undercomplete network. Also, we can see that the features in the  deep layers of an undercomplete network are more coarse when compared to overcomplete network. In fact the feature maps in the deep layers of an overcomplete network are more fine detailed due to the large feature size. These differences show the superiority of overcomplete representations from a convolutional neural network perspective.

\subsection{Network Architecture}
Now that we have established how an overcomplete deep network can be made to learn overcomplete latent representations of the data, we discuss how it can be designed to solve subspace clustering. 

In ODSC, we propose using two encoders which get trained in parallel. One is a generic encoder which has max-pooling layers after every convolution layer. Another is an overcomplete encoder where we have an upsampling layer after every convolution layer. For upsampling, after exploring both learned convolution and using a bilinear interpolation method, we found that both the methods resulted in equal performance. So, we chose to use bilinear interpolation for upsampling in our architecture for simplicity. In the latent space, we fuse both the latent representations before passing them to the self-expressive layer. The latent overcomplete representations are passed through a max-pooling layer before being fused with the latent representations of the undercomplete encoder. The reason behind this fusion approach is that even though overcomplete representations are better and more meaningful for clustering when compared to the undercomplete representations, they are relatively larger in the spatial sense and so we would need more parameters in our self-expressive layer to accomodate them. This makes the training of the network difficult as when we have more number of parameters we need a lot of data to prevent overfitting. So, using fusion we are able to maintain less number of parameters in the self-expressive layer as in DSC while also being able to take the advantages of overcomplete representations. We pass this latent space representation to the self-expressive layer.

\begin{figure*}[htbp!]
	\begin{center}
		\centering
		\includegraphics[width=0.8\textwidth]{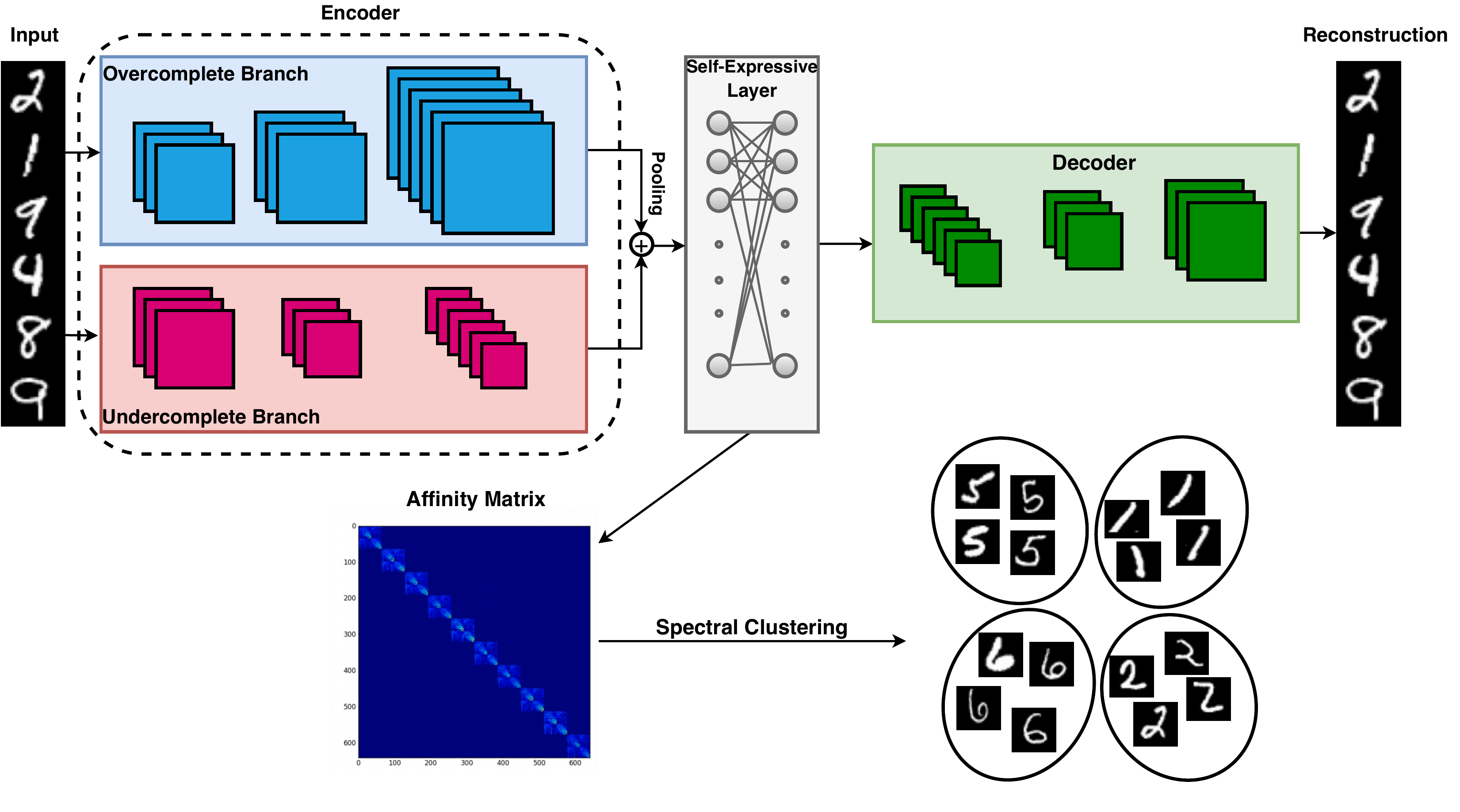}
		\vskip-10pt	\caption{Overall approach for the proposed ODSC method.}
		\label{Fig:odsc}
	\end{center}
\end{figure*}

The self-expressive layer is a fully-connected linear layer where the weights of it correspond to the coefficients in the self-expression representation matrix $C$. This layer learns the affinity matrix directly. For the decoder part, we have a common decoder that consists of convolutional layers followed by upsampling layers. We do not choose to use an overcomplete decoder here because it leads to more parameters and does not contribute much in the performance as the latent representations are what matters more for self-expressive layer. The number of convolutional layers we use in both the encoder and decoder varies with respect to the dataset. We decide on that based on the total number of data available in each dataset. Using the affinity matrix learned, we perform spectral clustering to get the clusters. Figure \ref{Fig:odsc} illustrates the proposed network architecture and the ODSC method.

\subsection{Training Details}
The autoencoder part of our network is initially trained separately for the task of reconstruction in an unsupervised way. It is trained with a reconstruction loss which is just the mean squared error (MSE) calculated between the reconstruction by the autoencoder $\hat{X}$ and the input image $X$. The reconstruction loss $\mathcal{L}_r$ is formulated as follows:
\begin{equation}
\mathcal{L}_r = \|X-\hat{X}\|_F^2. 
\end{equation}
We use Adam optimizer \cite{kingma2014adam} with a learning rate of 0.001 to train the reconstruction network for all the experiments. Note that the self-expressive layer is not trained in this part. For subspace clustering, we start by loading these pre-trained weights into the network. Then, we fine-tune the network by using the self-expressive layer and a self-expressive loss term $\mathcal{L}_{self}$, which can be formulated as follows:
\begin{equation}
\mathcal{L}_{self}(\theta,C) = \lambda_2 \|C\|_p + \frac{\lambda_3}{2} \|Z_{\theta_e} - Z_{\theta_e}C\|_F^2,
\end{equation}
where  $\theta$ represents the parameters of the network and $\theta_e$ specifically represents the parameters of the encoder.  $Z$ represents the latent representations found in the self-expressive layer of the network and $C$ represents the self-representation coefficient matrix. We use L2 regularization on C ($p=2$). For training the network in the fine-tuning stage, we optimize the loss using Adam optimizer on the combination of both of these losses. The final loss $\mathcal{L}_{Total}$ is defined as follows:
\[\mathcal{L}_{Total} = \frac{\lambda_1}{2} \mathcal{L}_r + \mathcal{L}_{self} \]
\begin{equation}	
\mathcal{L}_{Total} =  \frac{\lambda_1}{2} \|X-\hat{X}\|_F^2 + \lambda_2 \|C\|_p + \frac{\lambda_3}{2} \|Z_{\theta_e} - Z_{\theta_e}C\|_F^2,
\end{equation}
where $\lambda_1,\lambda_2$ and $\lambda_3$ are the hyperparameters that control the amount of effect each separate loss term can have over the total loss. These hyperparameter settings are discussed in the next section separately for each dataset. The whole training process is unsupervised as we do not make use of any labels.  We make use of only the input data and latent representations derived from the input for training our network. After fine-tuning, we use the parameters of the self-expressive layer to construct the affinity matrix for spectral clustering \cite{ng2002spectral}. We follow the same heuristics used by SSC \cite{elhamifar2013sparse} for this step.

\section{Experiments}
We perform all our experiments in Python using Tensorflow-1.14 \cite{abadi2016tensorflow} and evaluate our method using four datasets - MNIST \cite{lecun1998gradient}, COIL20 \cite{nene1996columbia}, ORL \cite{samaria1994parameterisation} and Extended Yale B \cite{lee2005acquiring}. Our method is compared with the following baselines: Low Rank Representation
(LRR) \cite{liu2012robust}, Low Rank Subspace Clustering (LRSC) \cite{vidal2014low},   Sparse Subspace Clustering (SSC) \cite{elhamifar2013sparse}, Kernel Sparse Subspace Clustering (KSSC) \cite{patel2014kernel}, SSC by Orthogonal Matching Pursuit (SSCOMP) \cite{you2016scalable}, Efficient Dense Subspace Clustering (EDSC) \cite{ji2014efficient}, SSC with the pre-trained convolutional auto-encoder features (AE+SSC), EDSC with the pre-trained convolutional auto-encoder features
(AE+EDSC) and Deep Subspace Clustering (DSC) Networks \cite{ji2017deep}. In the following sections, we will discuss in detail the hyperparameters we use for each dataset and the improvements we obtain over the existing methods.  The hyperparameters vary for each dataset because the number of data in each dataset varies. Note that for fair comparison, in all the experiments we either matched or used less number of parameters in the self-expressive layer when compared to DSC by lowering the number of filters used in a layer even though we use two encoders.  

\subsection{MNIST Dataset}
The MNIST dataset has a collection of hand-written digit images from 0 to 9. We randomly pick 100 images out of each class and use this collection of 1000 images for the task of subspace clustering. The size of the images is $28 \times 28$. It can be noted that the MNIST dataset accounts for many deformations caused by the style of hand-writing even among a single class making the task of clustering difficult as it is an unsupervised setting. The network architecture (ODSC) we use for this dataset has 2 convolutional blocks in the overcomplete branch of encoder, 3 in the undercomplete branch of encoder and 3 in the decoder. The details of what is present in each convolutional block were explained in the previous section. The kernel size of the convolutional layer is $5 \times 5$ in the first layer and $3 \times 3$ in every other layer in the encoder and vice- versa in the decoder. The number of filters is 20, 10 in the overcomplete branch of encoder and 20, 10, 5 in the undercomplete branch in order, respectively. In decoder, the number of filters are 5, 10, 20 in each convolutional block, respectively.  We set $\lambda_1 = 1.00, \lambda_2 = 20.00$ and $ \lambda_3 = 0.1$.  The network is fine-tuned for 100 epochs. The results in terms of clustering error are tabulated in Table \ref{mnist} where it can be seen that our ODSC method  achieves an improvement of 6.2 \% when compared to DSC. 

\begin{table*}[htp!]
	\begin{center}
		\centering

		\resizebox{1.\textwidth}{!}{
			
			\begin{tabular}{c|cccccccc|cc|cc}
				
				\hline
				Method       & SSC   & ENSC  & KSSC   & SSC-OMP   & EDSC  & LRR   & LRSC   & AE+SSC  & DSC-L1 & DSC-L2  &ODSC \\ \hline 
				Error       & 54.70  & 50.17 & 47.80 & 66.00   & 43.50  & 46.14  & 48.60  & 51.60  & 27.20 & 25.00    &  \textbf{18.80}    \\ \hline
				
			\end{tabular}
			
		}
	\end{center}
	\caption{Comparison of clustering error with recent methods for the MNIST dataset. }
	\label{mnist}
\end{table*}

\subsection{ORL Dataset}
The ORL dataset has a collection of face images corresponding to 40 subjects with 10 samples for each person. The whole collection of 400 images is used in the experiment for subspace clustering. The images are resized to $32 \times 32$. It can be noted that the ORL dataset consists of images under different lighting conditions and with different facial expressions. Also, this dataset is relatively smaller with just 400 images. The network architecture (ODSC) we use for this dataset has 2 convolutional blocks in the overcomplete branch of encoder, 3 in the undercomplete branch of encoder and 3 in the decoder. The kernel size of the convolutional layer is $3 \times 3$ for all the convolution layers in both encoder and decoder. The number of filters is 3, 3, and 6 for each block in encoder and in 6, 3 and 3 for each block in the decoder in the same order respectively. We set $\lambda_1 = 1.00, \lambda_2 = 2.00$ and $\lambda_3 = 0.1$ . The network is fine-tuned for 800 epochs.  The results in terms of clustering error are tabulated in Table \ref{orl} where it can be seen that our proposed method ODSC achieves an improvement of 2.00 \%  compared to DSC. 

\begin{table*}[htp!]
	\begin{center}
		\centering

		\resizebox{1.\textwidth}{!}{
			
			\begin{tabular}{c|cccccccc|cc|cc}
				
				\hline
				Method       & SSC   & ENSC  & KSSC   & SSC-OMP   & EDSC  & LRR   & LRSC   & AE+SSC  & DSC-L1 & DSC-L2  &ODSC \\ \hline 
				Error       & 32.50  & 24.75 & 34.25 & 36.00   & 27.25  & 38.25  & 32.5  & 26.75  & 14.25 & 14.00     &  \textbf{12.00}    \\ \hline

			\end{tabular}
			
		}
	\end{center}
	\caption{Comparison of clustering error with recent methods for the ORL dataset. }
	\label{orl}
\end{table*}

\subsection{COIL20 Dataset}
The COIL20 dataset has a collection of different object images. It consists of 20 classes and 1440 images. The images are resized to $32 \times 32$. The main challenge of this dataset is that different samples of the same object are captured in different angles which makes even the images of the same object look very different. The network architecture (ODSC) proposed  for this dataset has 1 convolutional block in both the undercomplete and overcomplete branches of encoder, 1 convolutional block in the decoder. The kernel size of the convolutional layer is $3 \times 3$ and the number of filters is 15 for all the convolution layers in both encoder and decoder.   We set $\lambda_1 = 1.00,\lambda_2=1.00$ and $ \lambda_3 = 15.00$. The network is fine-tuned for 40 epochs. The results in terms of clustering error for COIL20 are tabulated in Table \ref{coil20}.  Our proposed method ODSC achieves an improvement of 2.64 \% over DSC for COIL20 dataset.

\begin{table*}[htp!]
	\begin{center}
		\centering

		\resizebox{1.\textwidth}{!}{
			
			\begin{tabular}{c|cccccccc|cc|cc}
				
				\hline
				Method       & SSC   & ENSC  & KSSC   & SSC-OMP   & EDSC  & LRR   & LRSC   & AE+SSC  & DSC-L1 & DSC-L2  &ODSC \\ \hline 
				Error       & 14.86  & 12.40 & 24.65 & 45.90   & 14.86  & 31.01  & 31.25  & 22.08  & 6.95 & 5.14     &  \textbf{2.5}    \\ \hline

			\end{tabular}
			
		}
	\end{center}
	\caption{Comparison of clustering error with recent methods for the COIL20 dataset. }
	\label{coil20}
\end{table*}

\begin{table*}[htp!]
	\begin{center}
		\centering

		\resizebox{1.\textwidth}{!}{
			
			\begin{tabular}{c|cccccccc|cc|cc}
				
				\hline
				Method       & SSC   & ENSC  & KSSC   & SSC-OMP   & EDSC  & LRR   & LRSC   & AE+SSC  & DSC-L1 & DSC-L2  &ODSC \\ \hline 
				Error       & 27.51  & 12.40 & 27.75 & 24.71   & 11.64  & 34.81  & 29.89  & 25.33  & 3.33 & 2.67     &  \textbf{2.22 }    \\ \hline

			\end{tabular}
			
		}
	\end{center}
	\caption{Comparison of clustering error with recent methods for the Extended Yale B dataset. }
	\label{eyaleb}
\end{table*}






\subsection{Extended YaleB Dataset}
The Extended YaleB Dataset has a collection of face images which are taken under varying light conditions. It consists of 38 classes with 64 images per class resulting in a total of 2432 images. The images are resized to $48 \times 42$ for easy comparison with baselines. The network architecture (ODSC) we use for this dataset has 2 convolutional blocks in the overcomplete branch of encoder, 3 in the undercomplete branch of encoder and 3 in the decoder. The kernel size of the convolutional layer is $5 \times 5$ in the first layer and $3 \times 3$ in every other layer in the encoder and vice-versa in the decoder. The number of filters is 2 across all the convolutional layers in the dataset. We set $\lambda_1 = 1.00, \lambda_2 = 1.00$ and $\lambda_3 = 6.30 $.  The network is fine-tuned for 800 epochs. The results in terms of clustering error are tabulated in Table \ref{eyaleb}. We note here that even though ODSC achieves a better performance than other methods, this dataset has already reached saturation in terms of performance so our margin of improvement here is not as signifiacnt as that of the other datasets.

\section{Discussion}

\begin{table}[htp!]
	\begin{center}
		
		\resizebox{0.8\linewidth}{!}{
			
			\begin{tabular}{c|cccccccc|cc|cc}
				
				\hline
				Method       & DSC (U)   &  DSC (O)    & ODSC   \\ \hline 
				MNIST       & 25.00   & 21.60  & \textbf{18.8}        \\ \hline
				ORL       & 14.00   & 12.75  & \textbf{12.00}        \\ \hline
				COIL20       & 5.14   & 3.20  & \textbf{2.50}        \\ \hline
				
				EYaleB       & 2.67   & 2.35 & \textbf{2.22}          \\ \hline
				
			\end{tabular}
			
		}
		\vskip 6pt	\caption{Ablation study. The numbers correspond to the error in terms of percentage.}
		\label{abl}
	\end{center}
\end{table}

In this section, we first report a detailed ablation study and then discuss  other characteristics of the proposed method.

\subsection*{Ablation Study}
\begin{figure}[htp!]
	\centering
	
		\includegraphics[width=1\linewidth]{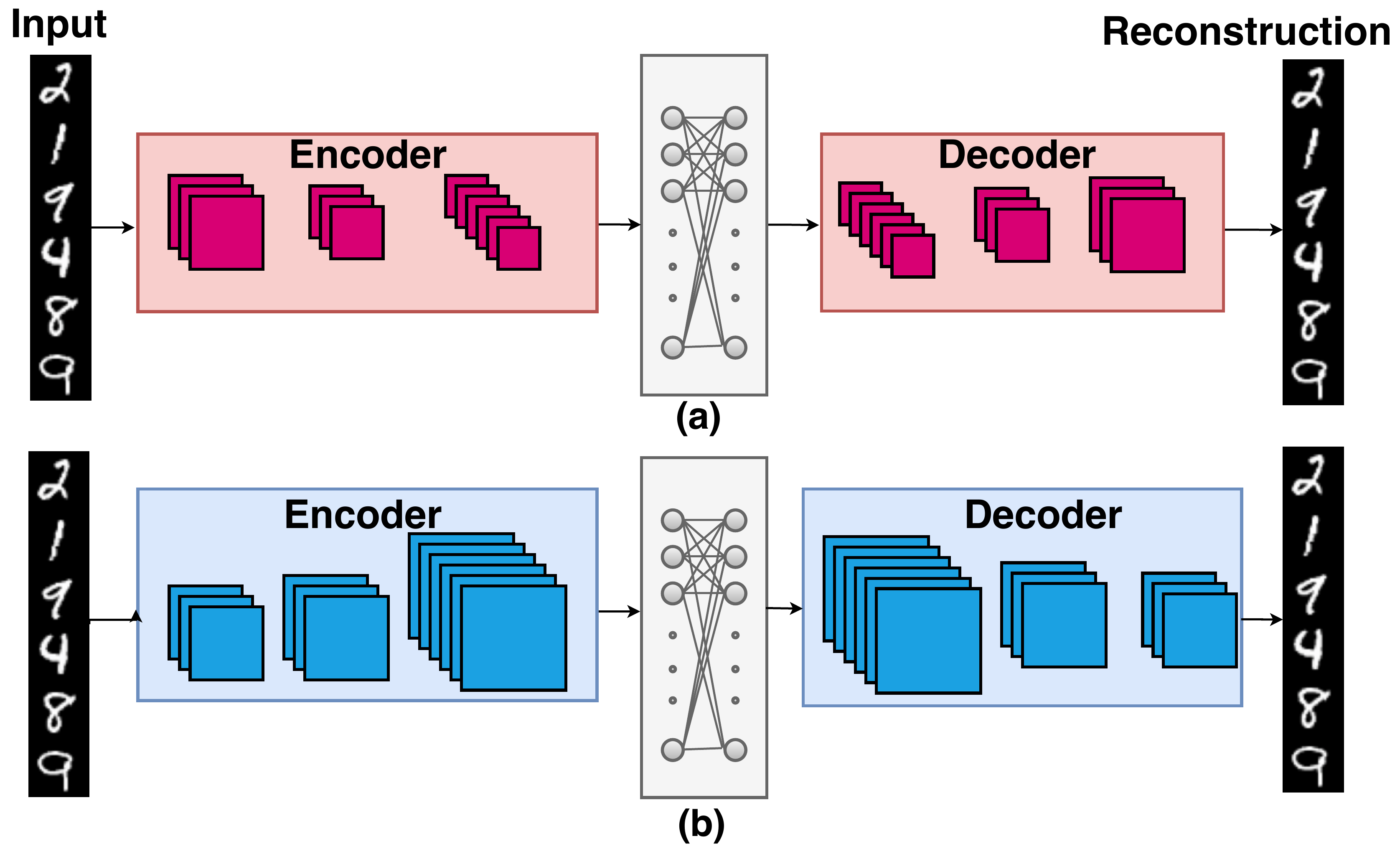}
		
		\caption{Architecture for ablation study: (a) DSC - Undercomplete (DSC-U). (b) DSC - Overcomplete (DSC-O).}
		\label{Fig:ab}
	
\end{figure}

For  ablation study, we start with DSC which uses an undercomplete deep auto-encoder. We represent this setup as DSC-U. Then, we change the deep-autoencoder architecture to overcomplete and represent this setup as DSC-O. The architectures of DSC-U and DSC-O are visualized in Figure \ref{Fig:ab}. We then show how our proposed method ODSC which has a fused encoder architecture of both undercomplete and overcomplete encoder fares compared to the other two. The results for all the four datsets can be found in Table \ref{abl}.  As can be seen from this table, in general DSC-O performs better than DSC-U on all datasets.  The best performance is obtained when both DSC-U and DSC-O are fused (i.e. ODSC).  This experiment clearly shows the significance of the proposed subspace clustering method.   


\subsection*{Reconstruction}
\begin{figure*}[htbp!]
	\begin{center}
		\centering
		\includegraphics[width=.8\textwidth]{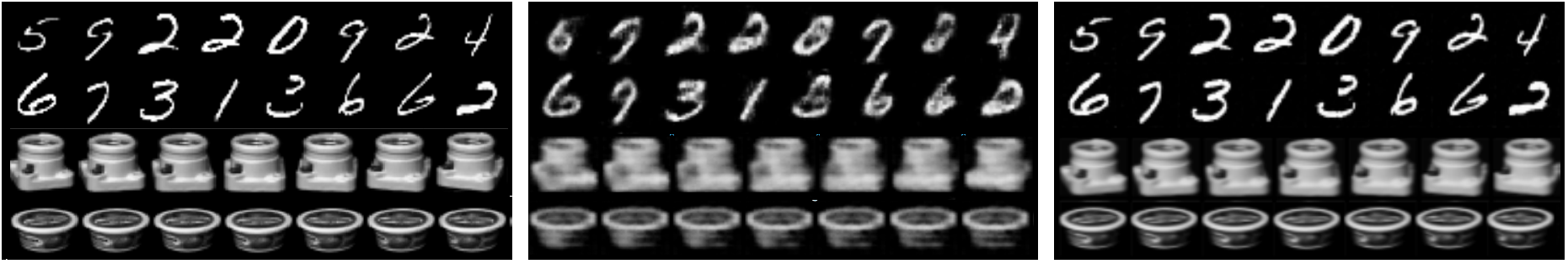}
		\\
		(a)\hskip130pt(b)\hskip125pt(c)\hskip110pt
		\caption{ (a) Input data. (b) Reconstructions using an  undercomplete deep auto-encoder. (c) Reconstructions using an overcomplete deep auto-encoder.}
		\label{Fig:recon}
	\end{center}
\end{figure*}

In both DSC and ODSC, the networks are first trained for reconstruction. Only from this process are the latent representations learned and are later fed into the self-expressive layer during fine-tuning. It is evident that better the latent representations, better the reconstructions. This is because only if the latent representations are meaningful, the decoder will be able to make a good reconstruction. In case of overcomplete architecture, as we project the images to a higher dimension, the feature maps learn very fine details. When visualized, these feature maps have a very good representation of the input image (please refer Fig \ref{Fig:feat}). Thus, for the overcomplete architecture we were able to get far better reconstruction when compared to undercomplete architecture. The reconstructions of both the types of architectures for the MNIST and COIL20 datasets are visualized in Figure \ref{Fig:recon}. From this figure, we can see that the overcomplete network is able to achieve better reconstructions and as a result the corresponding latent representations are more meaningful than undercomplete representations.   


\begin{table*}[htp!]
	\begin{center}
		\centering

		\resizebox{0.8\textwidth}{!}{
			
			\begin{tabular}{c|ccc|c|ccc|ccc|cc}
				
				\hline
				layer       & enc-1   & enc-2  & enc-3   & self-expressive & dec-1 & dec-2 &dec-3 & Total   \\ \hline 
				kernel size       & $5 \times 5$  & $3 \times 3$ & $3 \times 3$ & -   & $3 \times 3$  & $3 \times 3$  & $5 \times 5$  & -    \\ \hline
				channels       & 5  & 3 & 3 & -   & 3  & 3  & 5 & -     \\ \hline
				\# Parameters       & 130  & 138 & 84 & 160000   & 84  & 140  & 126 & 160702     \\ \hline
				
			\end{tabular}
			
		}
	\end{center}
	\caption{Details of the DSC network. }
	\label{dsc}
\end{table*}

\begin{table*}[htp!]
	\begin{center}
		\centering

		\resizebox{1.\textwidth}{!}{
			
			\begin{tabular}{c|ccccc|c|ccc|ccc}
				
				\hline
				layer       & enc-1 (U)  & enc-2 (U) & enc-3 (U) & enc-1 (O) & enc-2 (O)  & self-expressive & dec-1 & dec-2 &dec-3  & Total  \\ \hline 
				kernel size       & $3 \times 3$  & $3 \times 3$ & $3 \times 3$    & $3 \times 3$  & $3 \times 3$  & - & $3 \times 3$ & $3 \times 3$ & $3 \times 3$ & -     \\ \hline
				channels       & 3  & 3 & 3 & 3   & 3  & -  & 3 & 3 &3  & -    \\ \hline
				\# Parameters       & 30  & 84 & 84 & 30   & 84  & 160000  & 84 & 84 & 30 & 160510    \\ \hline
				
			\end{tabular}
			
		}
	\end{center}
	\caption{Details of the ODSC network. }
	\label{odsc}
\end{table*}

\subsection*{Robustness}

The performance of DSC significantly depends on the right place where the pretraining is stopped. The authors of DSC follow an approach of stopping the pretraining process when the reconstructions start to look reasonable. Stopping the training based on the reconstruction quality is not efficient. Moreover, stopping pretraining a few epochs before or after the desired instant leads to poor results in their case. A network with high dependence on this would be unstable and inefficient during real-world implementation. To this end, we show that  overcomplete representations are more robust and so do not depend much on things like where pretraining is stopped. Table \ref{pre} shows the results for an experiment we carried out where we stopped the pretraining at different epochs for both DSC and ODSC for the MNIST dataset. It can be observed that the change in the performance for ODSC is minimal when compared to that of DSC. 

In real-world, data is often noisy. Hence we compare the performance of different mehtods on noisy data.  We add random noise to the MNIST dataset and carry out clustering with both DSC and ODSC. The random noise is added in different levels so as to study the increase in error for the methods with respect to increase in noise. The results corresponding to this experiment are tabulated in Table \ref{mnist}. It can be seen that ODSC gives a better performance compared to DSC for all levels of noise. One major observation is that when the noise level is taken from $ 0\% $ to $ 50\% $, the error for DSC goes from 25.00 to 28.10, causing an increase of $3.10 \%$ error when the noise level is increased by $ 50\% $. However for the same increase in noise level, the increase in error for ODSC is just $ 1.10 \%$. This shows us that ODSC is more robust to addition of  noise when compared to DSC. 

\begin{table}[htp!]
	\begin{center}
		\centering

		\resizebox{0.5\textwidth}{!}{
			
			\begin{tabular}{c|cccccccc|cc|cc}
				
				\hline
				Method   &  0\%   & 10\%  & 20\%   & 30\%  & 40\%  & 50\%   \\ \hline 
				DSC    & 25.00  & 26.60  & 27.00 & 27.20  & 27.40 & 28.10       \\ \hline
				ODSC    & \textbf{18.80}  & \textbf{19.60}  & \textbf{19.70} & \textbf{20.00} & \textbf{19.90} & \textbf{19.90}       \\ \hline

			\end{tabular}
			
		}
	\end{center}
	\caption{Comparison of DSC and ODSC in terms of clustering error percentage for  MNIST data when added with different levels of noise. The first row corresponds to the amount of noise added to the data.  }
	\label{mnist}
\end{table}

\begin{table}[htp!]
	
\centering
		\resizebox{0.9\linewidth}{!}{
			\begin{tabular}{c|ccc|ccccc|cc|cc}
				
				\hline
				Pretraining Epoch       & 50    &  100   & 150 & Avg  \\ \hline 
				DSC       & 27.90    & 26.10  & 32.5  & 28.33 \\ \hline
				ODSC       & 21.50    & 18.80  & 21.80 & 20.7   \\ \hline

			\end{tabular}
		}	
		\vskip 6pt	\caption{Comparison of DSC and ODSC in terms of clustering error percentage while stopping pretraining at different epochs for MNIST dataset. }
		\label{pre}
		
\end{table}

\section*{Number of Parameters}

 Since ODSC has feature maps of a larger size when compared to DSC and also has two branches in the encoder, it will contain more number of parameters if we have the same number of filters as that of DSC. So, we reduce the number of filters across each layer in ODSC to match the number of parameters of DSC. We analyze this in detail for the architecture we used for ORL dataset as seen in Tables \ref{dsc} and \ref{odsc}. In Table \ref{odsc}, enc (U) corresponds to the undercomplete branch of encoder while enc (O) corresponds to the overcomplete branch of encoder. It can be seen that the number of parameters of ODSC is actually less than that of DSC. From this, we show that ODSC giving better results is not due to more number of parameters but due to the better properties of overcomplete representations. Also, as we make sure that the number of parameters in the self-expressive layer is the same for DSC and ODSC, so that the change in parameters is not that different for any experiment. For real-time  usage, we note that we can have more number of parameters for ODSC and might be able to get even better performance.

\section{Conclusion}
In this paper, we proposed a new solution to the problem of subspace clustering using overcomplete representations. Using a combination of overcomplete and undercomplete networks, we build an affinity matrix and perform spectral clustering to get the clusters. We performed extensive experiments on four benchmark datasets where our proposed method ODSC achieved better results. Also, we show that the latent representations learned by our method is more meaningful as we get better reconstructions and is also more robust to noise. ODSC is also not that dependent on pre-training as DSC. 


 \section*{Acknowledgment}
This work was supported by the NSF grant 1910141.

{\small
\bibliographystyle{ieee_fullname}
\bibliography{egbib}
}

\end{document}